\pdfoutput=1

\documentclass[11pt]{article}

\usepackage[]{ACL2023}

\usepackage{times}
\usepackage{natbib}
\usepackage{latexsym}
\usepackage{amsmath}
\usepackage{multirow}
\usepackage{tabularx} 
\usepackage{booktabs} 
\usepackage{adjustbox}
\usepackage[T1]{fontenc}

\usepackage[utf8]{inputenc}

\usepackage{microtype}

\usepackage{inconsolata}

\title{Strategic Data Ordering: Enhancing Large Language Model Performance through Curriculum Learning}

\author{Jisu Kim \\
  Diqest \\
  \texttt{jisukim8873@gmail.com} \\\And
  Juhwan Lee \\
  Diqest \\
  \texttt{9521ljh@gmail.com} \\}

\begin{document}
\maketitle
\begin{abstract}
The rapid advancement of Large Language Models (LLMs) has improved text understanding and generation but poses challenges in computational resources. This study proposes a curriculum learning-inspired, data-centric training strategy that begins with simpler tasks and progresses to more complex ones, using criteria such as prompt length, attention scores, and loss values to structure the training data. Experiments with Mistral-7B~\citep{jiang2023mistral} and Gemma-7B~\citep{team2024gemma} models demonstrate that curriculum learning slightly improves performance compared to traditional random data shuffling. Notably, we observed that sorting data based on our proposed attention criteria generally led to better performance. This approach offers a sustainable method to enhance LLM performance without increasing model size or dataset volume, addressing scalability challenges in LLM training.
\end{abstract}

\section{Introduction}

%
The rapid advancement of Large Language Models (LLMs) has been a defining trend in artificial intelligence, reshaping the capabilities of machines in understanding and generating human-like text. This surge in progress can be traced back to the introduction of the Transformer architecture~\citep{vaswani2017attention}, which laid the groundwork for the development of models such as the Generative Pre-trained Transformer (GPT) family. These models have been pivotal in pushing the boundaries of what machines can achieve, largely due to their expansive parameter counts and sophisticated training methodologies. The Transformer's introduction has not only catalyzed significant progress but also sparked a paradigm shift in how we approach machine learning tasks~\citep{acheampong2021transformer, patwardhan2023transformers, ali2024analyzing}. 

OpenAI's research underscores a key observation in the progression of these models: larger models tend to exhibit superior performance~\citep{kaplan2020scaling}. However, this approach of scaling up models introduces a set of challenges, notably the increased demand for computational resources and associated costs. As the size of models escalates, it prompts a crucial question about the sustainability of such growth trajectories and whether alternative strategies might offer more efficient pathways to improving model performance. 

In response to these challenges, we proposes a data-centric training strategy, inspired by the pedagogical concept of curriculum learning. Curriculum learning, a strategy that mimics human educational approaches by starting with simpler tasks and progressively moving to more complex ones. By structuring the training data and tasks in a manner that reflects the incremental learning stages humans go through, this method aims to facilitate a more natural and efficient learning process for the models. This approach deviates from traditional training practices, which often involve exposing models to a random assortment of data, suggesting a structured progression could yield better performance and efficiency. 

The potential of curriculum learning in the realm of LLMs is profound. It not only aligns with cognitive and educational theories but also presents a practical framework to tackle the scalability challenges associated with model training. By focusing on the complexity of training data and adopting a gradual learning, curriculum learning could lead to the better performance without the necessity for constant increases in model size. 

Therefore, we focuses on data-centric strategy training method by curriculumn learning, by examining the impact of performance of model, comparing random shuffled training dataset.

This study focuses on a data-centric strategy to enhance model performance through the adoption of curriculum learning. By examining the impact of structured training data, this study aims to demonstrate its effectiveness relative to the traditional method of random shuffling. We seek to establish that curriculum learning is a crucial strategy for improving the development process of large language models, potentially achieving better performance without increasing the model parameters or the size of the dataset.

\section{Related Work}
Curriculum learning, introduced by Bengio et al. in 2009~\citep{bengio2009curriculum}, is a learning strategy where models are trained on data from simple to complex, mirroring human learning processes. This approach enhances model performance by establishing a foundational understanding before introducing more complex data. It has been widely applied in natural language processing (NLP) and computer vision.~\citep{guo2018curriculumnet, antonios2019competence, tay2019simple}.

In NLP, curriculum learning has proven to be effective by organizing the training sequence to begin with simpler texts and progressively transition to more complex ones. The complexity of these texts is assessed based on various criteria, including text length, the rarity of vocabulary, and edit distance. This structured approach has successfully demonstrated the efficacy of curriculum learning in the training of Pretrained Language Models (PLMs)~\citep{chang2021does, nagatsuka2023length}.

Recent advancements in large-scale language and multimodal models have highlighted the effective integration of curriculum learning with cutting-edge techniques, significantly boosting performance across various applications. Notably, projects such as WavLLM~\citep{hu2024wavllm} and AutoWebGLM~\citep{lai2024autowebglm} showcase the potential of these integrations, tailoring the learning process to gradually introduce more complex tasks, thereby enhancing model robustness and application specificity. Alongside these innovations, further research has validated the effectiveness of curriculum learning in instruction tuning for language models. This approach not only systematically improves models' abilities to handle progressively challenging tasks but also optimizes their learning trajectory for better performance in specialized tasks. By methodically structuring the training process, these initiatives have set a new standard in the development and refinement of AI models, marking a significant step forward in their evolution.

In the development of training data for curriculum learning, accurately measuring data difficulty is crucial. Specifically, for LLMs (Large Language Models), determining the degree of data difficulty is challenging. Traditional metrics, such as text length or word rarity, are often employed to estimate the difficulty of training data~\citep{chang2021does, nagatsuka2023length}. However, these metrics may not fully reflect the complexity of a dataset. It is essential to assess data difficulty from the model's perspective, rather than relying solely on data-specific metrics. Our research proposes a new approach to calculate the degree of difficulty based on a model-centric perspective. By organizing the training dataset according to difficulty using our new metric, we aim to improve the model's performance compared to random shuffling.

\section{Methods for Quantitative Difficulty Measurement}
In this section, we introduce a novel methodology for training models that begins with easier tasks and methodically progresses to more challenging ones. This approach requires organizing data by its level of complexity, for which we have established three principal criteria. By arranging the data in an order that goes from less to more difficult, this approach establishes a structured progression for learning.

\subsection*{Length}
It is widely acknowledged in the study that longer sentences are more challenging to learn. To assess difficulty based on length, we address the measurement of length in tokenized prompt, a critical factor in understanding the complexities of language processing. The prompt length, denoted as $d_{\text{length}}(p)$, is defined as the total number of tokens produced by the tokenizer for a given prompt. Mathematically, it is represented as:
\begin{equation}
d_{\text{length}}(p) = N.
\end{equation}

\subsection*{Attention Score} 
We introduce a novel method to determine the complexity of a prompt by analyzing model attention scores. In our approach, we suggest that how a model focuses on different parts of a prompt—reflected by attention scores—can show how difficult the prompt is. If a model concentrates its attention on just a few token parts, the prompt is likely easier since only those parts are needed to find the answer. However, if the attention is spread evenly across the whole prompt, it's harder, indicating that understanding everything is necessary to get the right answer. We measure this by looking at the variation in attention scores, using two specific methods.

This study focus exclusively on the attention scores for tokens following the section where the response is generated to analyze response generation. We calculate the variance of attention scores within each attention layer for these tokens and then average these variances across all layers. This process is mathematically represented as:

\begin{enumerate}
    \item For each layer \(i\), calculate the variance of attention scores, \( \text{Var}(A_i) \), for tokens after the section where the response is generated.
    \item Compute the average variance across all \(N\) layers:
    \[ d_{\text{Average Variance}} = \frac{1}{N} \sum_{i=1}^{N} \text{Var}(A_i) \]
\end{enumerate}

\subsection*{Loss}
A high loss signifies a substantial discrepancy between the predicted and actual outcomes, indicating that the model perceives the data as challenging. We establish a direct correlation between high loss values and increased difficulty. Therefore, prompts that generate higher loss are considered more challenging for the model to learn.

This method computes the cross-entropy loss associated with the answer segment of the prompt.

\begin{equation}
d_{\text{loss}} = -\sum_{i} Prompt_i \log(Answer_i)
\end{equation}

\section{Experiment Settings}
\subsection*{Models}
In order to validate our hypotheses, this study conducted a series of experiments utilizing language models with 7 billion parameters. Specifically, we utilized the Mistral-7B and Gemma-7B models to conduct our analyses ~\citep{jiang2023mistral, team2024gemma}. 

In addition, because of limited resources, we utilized Quantize Low Rank Approximation (QLoRA)~\cite{hu2021lora} with Parameter-Efficient Fine-Tuning (PEFT) library to fine-tune the models. Different from full fine-tuning, LoRA freezes the pre-trained model weights and injects learnable rank decomposition matrices into each layer of the Transformer architecture. Thus, this method can reduce the number of trainable parameters, the cost of training and time for downstream tasks. The details of configurations are described on Table~\ref{tab:parameters}.

\begin{table*}[]
\centering
\caption{LoRA and Model hyperparameter settings.}
\label{tab:parameters}
\begin{tabular}{|l|llll|ll|}
\hline
           & \multicolumn{4}{c|}{LoRA parameters}                                                                                                                                                                    & \multicolumn{2}{c|}{Model parameters}            \\ \hline
Model      & \multicolumn{1}{l|}{Target Modules}                                                                                                   & \multicolumn{1}{l|}{Rank} & \multicolumn{1}{l|}{Alpa} & Dropout & \multicolumn{1}{l|}{Sliding window} & Rope theta \\ \hline
Mistral-7B & \multicolumn{1}{l|}{\begin{tabular}[c]{@{}l@{}}q\_proj, k\_proj, v\_proj, o\_proj, \\  gate\_proj, up\_proj, down\_proj\end{tabular}} & \multicolumn{1}{l|}{64}   & \multicolumn{1}{l|}{16}   & 0.1     & \multicolumn{1}{l|}{None}           & 1,000,000    \\ \hline
Gemma-7B   & \multicolumn{1}{l|}{\begin{tabular}[c]{@{}l@{}}q\_proj, k\_proj, v\_proj, o\_proj, \\  gate\_proj, up\_proj, down\_proj\end{tabular}} & \multicolumn{1}{l|}{64}   & \multicolumn{1}{l|}{16}   & 0.1     & \multicolumn{1}{l|}{-}              & 10,000          \\ \hline
\end{tabular}
\end{table*}

\subsection*{Instruction tuning Datasets}
This study focus on specific datasets that are considered the state-of-art results for instruction tuning. Due to the memory limitations of our hardware setup, this study is limited to using single-turn datasets.

\begin{itemize}
    \item \textbf{Orca-math}~\citep{mitra2024orca} This dataset contains answers generated for elementary school math word problems using Azure GPT-4 Turbo.
    \item \textbf{Alpaca}~\citep{alpaca} This dataset was created using the Self-Instruct~\citep{wang2022self} method and the OpenAI Text-Davinci-003 model.
    \item \textbf{SlimOrca-Dedup}~\citep{SlimOrcaDedup} This is a subset of the OpenOrca~\citep{mukherjee2023orca} data, where, within the FLAN~\citep{longpre2023flan} dataset, incorrect answers were identified and removed using OpenAI GPT-4 based on human annotations, and duplicates were eliminated in the SlimOrca~\citep{SlimOrca} dataset.
\end{itemize}

\section{Evaluation}

\subsection*{Language Model Evaluation Harness}
In order to evaluate instruction-tuned model, we used Language Model Evaluation Harness ~\citep{eval-harness} a framework for few-shot evaluation of language models. This method conducted ARC with 25 shot~\citep{clark2018think}, HellaSwag with 10 shots~\citep{zellers2019hellaswag}, MMLU with 5 shot~\citep{hendryckstest2021}, TrustfulQA with a 0 shot~\citep{lin2021truthfulqa}, Winogrande with 5 shot~\citep{ai2:winogrande} and GSM8K with 5 shot~\citep{cobbe2021gsm8k} datasets. The combination of these datasets extensively evaluates the ability of Large Language Models to respond to realistic questions and reasoning. By evaluating these datasets, we verify whether our proposed curriculum learning method impacts the model outcomes.

\section{Results}
\subsection*{Random Sorting in First Epoch with Subsequent Difficulty-Based Curriculum Learning}
To enhance the efficiency of training, we customized the HuggingFace Trainer to perform instruction tuning randomly during the initial epoch. From the second epoch, we systematically arranged the data to align with specific methods for instruction tuning. This approach was designed to assess the impact of structured versus random data presentation on the model's learning dynamics and overall performance.

In case of Mistral-7B, as shown in Table~\ref{tab:mistral7Balpaca}, instruction-tuning on the alpaca dataset did not yield performance improvements over the original model configuration. When comparing various instruction-tuning strategies, the most effective approach involved random data arrangement trained for only one epoch, achieving an accuracy of 60.91\%.

However, experiments conducted on the Orca-math dataset revealed that the most effective strategy involved aligning data based on attention mechanisms with three epochs. This approach resulted in the highest mean accuracy of 66.28\% as shown in Table~\ref{tab:mistral7Bmath}. 

When experimenting with the Slimorca dataset, the most effective training strategy involved organizing the data based on sequence length and limiting the training to two epochs. This method achieved the best performance with an accuracy of 65.08\% (Table~\ref{tab:mistral7Bslim}). 

For the Gemma-7B model, experiments conducted using the Alpaca dataset indicated that random data arrangement during instruction-tuning yielded the highest average performance, achieving an accuracy of 64.10\% (Tabel~\ref{tab:gemma7Balpaca}). 

Conversely, results from Table~\ref{tab:gemma7Bmath} and \ref{tab:gemma7Bslim}, conducted on the orca-math and slimorca-dedup datasets, indicate that attention-based curriculum learning was most effective. Specifically, in the orcamath dataset, this method achieved the best performance with an accuracy of 67.54\% after two epochs of training. Similarly, for the slimorca-dedup dataset, curriculum learning based on attention mechanisms and extending training to three epochs yielded the highest accuracy of 66.87\%.

\begin{table*}[ht]
\centering
\caption{Results of instruction-tuning the \textbf{Mistral-7B} model with the \textbf{Alpaca} dataset.}
\label{tab:mistral7Balpaca}
\begin{adjustbox}{scale=0.8} 
\begin{tabular}{|ll|l|l|l|l|l|l|l|}
\hline
\multicolumn{1}{|l|}{Epochs} & \multicolumn{1}{l|}{Methods} & \multicolumn{1}{l|}{Average (\%)} & \multicolumn{1}{l|}{\begin{tabular}[c]{@{}l@{}}ARC (\%) \\ (norm)\end{tabular}} & \multicolumn{1}{l|}{\begin{tabular}[c]{@{}l@{}}HellaSwag (\%)\\ (norm)\end{tabular}} & \multicolumn{1}{l|}{MMLU (\%)} & \multicolumn{1}{l|}{\begin{tabular}[c]{@{}l@{}}TrustfulQA (\%)\\ (mc2)\end{tabular}} & \multicolumn{1}{l|}{Winogrande (\%)} & \multicolumn{1}{l|}{\begin{tabular}[c]{@{}l@{}}GSM8K (\%)\\ (strict-match)\end{tabular}} \\ \hline
\multicolumn{1}{|l|}{1}                  & random    & \textbf{60.91}   & 62.12      & 82.81            & 61.37 & 43.74           & 79.24      & 36.16               \\ \hline
\multicolumn{1}{|l|}{\multirow{4}{*}{2}} & random    & 60.27   & 63.65      & 83.33            & 60.48 & 42.80           & 78.45      & 32.90               \\ \cline{2-9} 
\multicolumn{1}{|l|}{}                   & attention & 60.72   & 63.14      & 83.67            & 60.90 & 42.77           & 79.08      & 34.72               \\ \cline{2-9} 
\multicolumn{1}{|l|}{}                   & loss      & 59.98   & 63.40      & 83.31            & 60.65 & 45.79           & 78.53      & 28.20               \\ \cline{2-9} 
\multicolumn{1}{|l|}{}                   & length    & 60.58   & 62.88      & 83.64            & 61.01 & 42.90           & 79.24      & 33.81               \\ \hline
\multicolumn{1}{|l|}{\multirow{4}{*}{3}} & random    & 59.87   & 63.14      & 84.15            & 59.88 & 43.26           & 77.51      & 31.31               \\ \cline{2-9} 
\multicolumn{1}{|l|}{}                   & attention & 60.54   & 63.05      & 84.06            & 60.46 & 42.90           & 77.98      & 34.80               \\ \cline{2-9} 
\multicolumn{1}{|l|}{}                   & loss      & 59.84   & 64.93      & 84.15            & 59.54 & 45.84           & 77.98      & 26.61               \\ \cline{2-9} 
\multicolumn{1}{|l|}{}                   & length    & 59.95   & 63.82      & 84.21            & 60.38 & 43.33           & 76.48      & 31.46               \\ \hline
\multicolumn{2}{|l|}{Base Model}                     & 60.99   & 61.51      & 83.47            & 62.39 & 42.60           & 78.05      & 37.90               \\ \hline
\end{tabular}
\end{adjustbox}
\end{table*}

\begin{table*}[ht]
\centering
\caption{Results of instruction-tuning the \textbf{Mistral-7B} model with the \textbf{Orca-math} dataset.}
\label{tab:mistral7Bmath}
\begin{adjustbox}{scale=0.8} 
\begin{tabular}{|l|l|l|l|l|l|l|l|l|}
\hline
\multicolumn{1}{|l|}{Epochs} & \multicolumn{1}{l|}{Methods} & \multicolumn{1}{l|}{Average (\%)} & \multicolumn{1}{l|}{\begin{tabular}[c]{@{}l@{}}ARC (\%) \\ (norm)\end{tabular}} & \multicolumn{1}{l|}{\begin{tabular}[c]{@{}l@{}}HellaSwag (\%)\\ (norm)\end{tabular}} & \multicolumn{1}{l|}{MMLU (\%)} & \multicolumn{1}{l|}{\begin{tabular}[c]{@{}l@{}}TrustfulQA (\%)\\ (mc2)\end{tabular}} & \multicolumn{1}{l|}{Winogrande (\%)} & \multicolumn{1}{l|}{\begin{tabular}[c]{@{}l@{}}GSM8K (\%)\\ (strict-match)\end{tabular}} \\ \hline
\multicolumn{1}{|l|}{1}                  & random    & 65.51   & 61.01      & 82.94            & 61.19 & 46.36           & 77.03      & 64.52               \\ \hline
\multicolumn{1}{|l|}{\multirow{4}{*}{2}} & random    & 66.16   & 60.24      & 82.92            & 60.23 & 47.45           & 76.87      & 69.22               \\ \cline{2-9} 
\multicolumn{1}{|l|}{}                   & attention & 66.20   & 61.77      & 83.08            & 59.66 & 45.57           & 77.90      & 69.22               \\ \cline{2-9} 
\multicolumn{1}{|l|}{}                   & loss      & 65.40   & 61.18      & 83.04            & 60.63 & 46.73           & 77.51      & 63.31               \\ \cline{2-9} 
\multicolumn{1}{|l|}{}                   & length    & 65.89   & 60.32      & 82.92            & 60.33 & 46.17           & 77.90      & 67.70               \\ \hline
\multicolumn{1}{|l|}{\multirow{4}{*}{3}} & random    & 65.95   & 60.49      & 82.86            & 59.64 & 48.69           & 76.24      & 67.78               \\ \cline{2-9} 
\multicolumn{1}{|l|}{}                   & attention & \textbf{66.28}   & 60.15      & 83.35            & 59.27 & 47.15           & 78.37      & 69.37               \\ \cline{2-9} 
\multicolumn{1}{|l|}{}                   & loss      & 65.58   & 60.32      & 82.79            & 59.60 & 47.57           & 75.93      & 67.25               \\ \cline{2-9} 
\multicolumn{1}{|l|}{}                   & length    & 65.61   & 60.92      & 83.16            & 58.92 & 47.76           & 77.03      & 65.88               \\ \hline
\multicolumn{2}{|l|}{Base Model}                     & 60.99   & 61.51      & 83.47            & 62.39 & 42.60           & 78.05      & 37.90               \\ \hline
\end{tabular}
\end{adjustbox}
\end{table*}

\begin{table*}[ht]
\centering
\caption{Results of instruction-tuning the \textbf{Mistral-7B} model with the \textbf{SlimOrca-Dedup} dataset.}
\label{tab:mistral7Bslim}
\begin{adjustbox}{scale=0.8} 
\begin{tabular}{|ll|l|l|l|l|l|l|l|}
\hline
\multicolumn{1}{|l|}{Epochs} & \multicolumn{1}{l|}{Methods} & \multicolumn{1}{l|}{Average (\%)} & \multicolumn{1}{l|}{\begin{tabular}[c]{@{}l@{}}ARC (\%) \\ (norm)\end{tabular}} & \multicolumn{1}{l|}{\begin{tabular}[c]{@{}l@{}}HellaSwag (\%)\\ (norm)\end{tabular}} & \multicolumn{1}{l|}{MMLU (\%)} & \multicolumn{1}{l|}{\begin{tabular}[c]{@{}l@{}}TrustfulQA (\%)\\ (mc2)\end{tabular}} & \multicolumn{1}{l|}{Winogrande (\%)} & \multicolumn{1}{l|}{\begin{tabular}[c]{@{}l@{}}GSM8K (\%)\\ (strict-match)\end{tabular}} \\ \hline
\multicolumn{1}{|l|}{1}                  & random    & 65.04   & 63.14      & 83.09            & 62.43 & 49.68           & 79.64      & 52.24               \\ \hline
\multicolumn{1}{|l|}{\multirow{4}{*}{2}} & random    & 64.53   & 59.90      & 83.51            & 60.95 & 50.87           & 78.22      & 53.75               \\ \cline{2-9} 
\multicolumn{1}{|l|}{}                   & attention & 64.67   & 62.03      & 83.35            & 60.95 & 49.58           & 77.90      & 54.21               \\ \cline{2-9} 
\multicolumn{1}{|l|}{}                   & loss      & 63.94   & 63.13      & 83.27            & 59.20 & 51.89           & 78.61      & 47.53               \\ \cline{2-9} 
\multicolumn{1}{|l|}{}                   & length    & \textbf{65.08}   & 62.12      & 83.41            & 61.76 & 50.31           & 78.93      & 53.98               \\ \hline
\multicolumn{1}{|l|}{\multirow{4}{*}{3}} & random    & 64.21   & 62.46      & 82.95            & 59.71 & 50.90           & 76.40      & 52.84               \\ \cline{2-9} 
\multicolumn{1}{|l|}{}                   & attention & 64.22   & 60.32      & 83.24            & 59.60 & 51.46           & 77.27      & 53.45               \\ \cline{2-9} 
\multicolumn{1}{|l|}{}                   & loss      & 65.01   & 65.35      & 82.95            & 60.09 & 53.63           & 78.29      & 49.73               \\ \cline{2-9} 
\multicolumn{1}{|l|}{}                   & length    & 64.40   & 61.60      & 83.64            & 60.20 & 51.55           & 76.48      & 52.92               \\ \hline
\multicolumn{2}{|l|}{Base Model}                     & 60.99   & 61.51      & 83.47            & 62.39 & 42.60           & 78.05      & 37.90               \\ \hline
\end{tabular}
\end{adjustbox}
\end{table*}

\begin{table*}[ht]
\centering
\caption{Results of instruction-tuning the \textbf{Gemma-7B} model with the \textbf{Alpaca} dataset.}
\label{tab:gemma7Balpaca}
\begin{adjustbox}{scale=0.8} 
\begin{tabular}{|ll|l|l|l|l|l|l|l|}
\hline
\multicolumn{1}{|l|}{Epochs} & \multicolumn{1}{l|}{Methods} & \multicolumn{1}{l|}{Average (\%)} & \multicolumn{1}{l|}{\begin{tabular}[c]{@{}l@{}}ARC (\%) \\ (norm)\end{tabular}} & \multicolumn{1}{l|}{\begin{tabular}[c]{@{}l@{}}HellaSwag (\%)\\ (norm)\end{tabular}} & \multicolumn{1}{l|}{MMLU (\%)} & \multicolumn{1}{l|}{\begin{tabular}[c]{@{}l@{}}TrustfulQA (\%)\\ (mc2)\end{tabular}} & \multicolumn{1}{l|}{Winogrande (\%)} & \multicolumn{1}{l|}{\begin{tabular}[c]{@{}l@{}}GSM8K (\%)\\ (strict-match)\end{tabular}} \\ \hline
\multicolumn{1}{|l|}{1}                  & random    & 62.70   & 61.00      & 81.95            & 62.43 & 47.25           & 77.58      & 46.01               \\ \hline
\multicolumn{1}{|l|}{\multirow{4}{*}{2}} & random    & \textbf{64.10}   & 62.96      & 82.32            & 64.20 & 48.63           & 77.66      & 48.82               \\ \cline{2-9} 
\multicolumn{1}{|l|}{}                   & attention & 63.94   & 61.86      & 82.32            & 63.20 & 50.40           & 77.27      & 48.60               \\ \cline{2-9} 
\multicolumn{1}{|l|}{}                   & loss      & 62.55   & 61.26      & 81.97            & 61.80 & 50.65           & 77.19      & 42.45               \\ \cline{2-9} 
\multicolumn{1}{|l|}{}                   & length    & 62.40   & 62.37      & 82.31            & 61.77 & 47.85           & 77.11      & 42.98               \\ \hline
\multicolumn{1}{|l|}{\multirow{4}{*}{3}} & random    & \textbf{64.10}   & 63.05      & 82.28            & 62.60 & 51.74           & 77.03      & 47.91               \\ \cline{2-9} 
\multicolumn{1}{|l|}{}                   & attention & 64.06   & 61.94      & 82.96            & 62.95 & 49.44           & 77.34      & 49.73               \\ \cline{2-9} 
\multicolumn{1}{|l|}{}                   & loss      & 62.66   & 59.98      & 82.28            & 62.71 & 52.61           & 76.47      & 41.92               \\ \cline{2-9} 
\multicolumn{1}{|l|}{}                   & length    & 63.92   & 61.95      & 82.96            & 62.95 & 49.45           & 77.35      & 48.89               \\ \hline
\multicolumn{2}{|l|}{Base Model}                     & 63.30   & 61.00      & 82.54            & 62.82 & 45.07           & 77.74      & 50.64               \\ \hline
\end{tabular}
\end{adjustbox}
\end{table*}

\begin{table*}[ht]
\centering
\caption{Results of instruction-tuning the \textbf{Gemma-7B} model with the \textbf{Orca-math} dataset.}
\label{tab:gemma7Bmath}
\begin{adjustbox}{scale=0.8} 
\begin{tabular}{|ll|l|l|l|l|l|l|l|}
\hline
\multicolumn{1}{|l|}{Epochs} & \multicolumn{1}{l|}{Methods} & \multicolumn{1}{l|}{Average (\%)} & \multicolumn{1}{l|}{\begin{tabular}[c]{@{}l@{}}ARC (\%) \\ (norm)\end{tabular}} & \multicolumn{1}{l|}{\begin{tabular}[c]{@{}l@{}}HellaSwag (\%)\\ (norm)\end{tabular}} & \multicolumn{1}{l|}{MMLU (\%)} & \multicolumn{1}{l|}{\begin{tabular}[c]{@{}l@{}}TrustfulQA (\%)\\ (mc2)\end{tabular}} & \multicolumn{1}{l|}{Winogrande (\%)} & \multicolumn{1}{l|}{\begin{tabular}[c]{@{}l@{}}GSM8K (\%)\\ (strict-match)\end{tabular}} \\ \hline
\multicolumn{1}{|l|}{1}                  & random    & 66.80   & 60.84      & 81.71            & 61.51 & 52.21           & 77.11      & 67.40               \\ \hline
\multicolumn{1}{|l|}{\multirow{4}{*}{2}} & random    & 66.95   & 61.35      & 81.89            & 60.52 & 53.44           & 75.22      & 69.29               \\ \cline{2-9} 
\multicolumn{1}{|l|}{}                   & attention & \textbf{67.54}   & 62.37      & 81.93            & 60.66 & 55.83           & 76.87      & 67.55               \\ \cline{2-9} 
\multicolumn{1}{|l|}{}                   & loss      & 67.01   & 61.43      & 82.11            & 61.00 & 55.23           & 76.95      & 65.35               \\ \cline{2-9} 
\multicolumn{1}{|l|}{}                   & length    & 66.87   & 61.86      & 81.64            & 60.10 & 53.86           & 76.09      & 67.70               \\ \hline
\multicolumn{1}{|l|}{\multirow{4}{*}{3}} & random    & 67.09   & 61.35      & 81.94            & 60.56 & 53.09           & 75.45      & 70.13               \\ \cline{2-9} 
\multicolumn{1}{|l|}{}                   & attention & 66.64   & 62.20      & 82.03            & 60.18 & 54.74           & 76.87      & 63.84               \\ \cline{2-9} 
\multicolumn{1}{|l|}{}                   & loss      & 66.96   & 61.69      & 82.08            & 60.63 & 55.42           & 77.27      & 64.67               \\ \cline{2-9} 
\multicolumn{1}{|l|}{}                   & length    & 66.25   & 62.12      & 81.66            & 60.52 & 54.37           & 75.93      & 62.93               \\ \hline
\multicolumn{2}{|l|}{Base Model}                     & 63.30   & 61.00      & 82.54            & 62.82 & 45.07           & 77.74      & 50.64               \\ \hline
\end{tabular}
\end{adjustbox}
\end{table*}

\begin{table*}[ht]
\centering
\caption{Results of instruction-tuning the \textbf{Gemma-7B} model with the \textbf{SlimOrca-Dedup} dataset.}
\label{tab:gemma7Bslim}
\begin{adjustbox}{scale=0.8} 
\begin{tabular}{|ll|l|l|l|l|l|l|l|}
\hline
\multicolumn{1}{|l|}{Epochs} & \multicolumn{1}{l|}{Methods} & \multicolumn{1}{l|}{Average (\%)} & \multicolumn{1}{l|}{\begin{tabular}[c]{@{}l@{}}ARC (\%) \\ (norm)\end{tabular}} & \multicolumn{1}{l|}{\begin{tabular}[c]{@{}l@{}}HellaSwag (\%)\\ (norm)\end{tabular}} & \multicolumn{1}{l|}{MMLU (\%)} & \multicolumn{1}{l|}{\begin{tabular}[c]{@{}l@{}}TrustfulQA (\%)\\ (mc2)\end{tabular}} & \multicolumn{1}{l|}{Winogrande (\%)} & \multicolumn{1}{l|}{\begin{tabular}[c]{@{}l@{}}GSM8K (\%)\\ (strict-match)\end{tabular}} \\ \hline
\multicolumn{1}{|l|}{1}                  & random    & 66.49   & 62.97      & 81.85            & 61.76 & 54.40           & 76.48      & 61.49               \\ \hline
\multicolumn{1}{|l|}{\multirow{4}{*}{2}} & random    & 66.50   & 64.25      & 82.07            & 60.75 & 53.31           & 75.85      & 62.77               \\ \cline{2-9} 
\multicolumn{1}{|l|}{}                   & attention & 66.73   & 64.08      & 81.37            & 62.11 & 53.24           & 74.82      & 64.75               \\ \cline{2-9} 
\multicolumn{1}{|l|}{}                   & loss      & 64.89   & 63.40      & 81.43            & 61.27 & 51.90           & 75.37      & 55.95               \\ \cline{2-9} 
\multicolumn{1}{|l|}{}                   & length    & 66.79   & 64.16      & 81.89            & 62.26 & 53.63           & 76.56      & 62.24               \\ \hline
\multicolumn{1}{|l|}{\multirow{4}{*}{3}} & random    & 66.36   & 63.14      & 81.90            & 61.19 & 55.88           & 74.66      & 61.41               \\ \cline{2-9} 
\multicolumn{1}{|l|}{}                   & attention & \textbf{66.87}   & 63.65      & 81.59            & 62.30 & 54.98           & 76.56      & 62.17               \\ \cline{2-9} 
\multicolumn{1}{|l|}{}                   & loss      & 65.48   & 64.08      & 81.30            & 61.87 & 52.93           & 75.85      & 56.86               \\ \cline{2-9} 
\multicolumn{1}{|l|}{}                   & length    & 66.55   & 62.54      & 81.46            & 61.80 & 55.98           & 74.98      & 62.55               \\ \hline
\multicolumn{2}{|l|}{Base Model}                     & 63.30   & 61.00      & 82.54            & 62.82 & 45.07           & 77.74      & 50.64               \\ \hline
\end{tabular}
\end{adjustbox}
\end{table*}

\section{Conclusion and Future Works}
While our proposed attention-based method has shown a tendency to improve instruction tuning, the differences observed have not been significantly impactful. Consequently, we plan to upload the results evaluated using the alpacaeval dataset in version 2 of our project. Currently, our instruction tuning approach involves random training for the first epoch, followed by structured data training in subsequent epochs. Therefore, the results are heavily reliant on the outcomes from the randomly trained model. In the next version, we intend to prepare a pre-arranged dataset and begin the training from the first epoch with this structured data to better assess the effectiveness of this approach.
\bibliography{anthology,custom}
\bibliographystyle{acl_natbib}




\end{document}